%% file: main.tex
\documentclass[10pt,twocolumn,letterpaper]{article}

\usepackage{cvpr}              

\usepackage{booktabs} 
\usepackage{tabularx}
\usepackage{arydshln}
\input{preamble}

\definecolor{cvprblue}{rgb}{0.21,0.49,0.74}
\usepackage[pagebackref,breaklinks,colorlinks,citecolor=cvprblue]{hyperref}

\def\paperID{19} 
\def\confName{CVPR}
\def\confYear{2024}

\title{Food Portion Estimation via 3D Object Scaling}

\author{
Gautham Vinod\quad
Jiangpeng He\quad
Zeman Shao\quad
Fengqing Zhu
\\
Purdue University, West Lafayette, Indiana, U.S.A. \\
{\tt\small \{gvinod, he416, shao112, zhu0\}@purdue.edu}}

\begin{document}
\maketitle
\input{sec/0_abstract}    
\input{sec/1_intro}
\input{sec/2_related_works}
\input{sec/3_method}
\input{sec/4_experimental_results}
\input{sec/5_conclusion}
{
    \small
    \bibliographystyle{ieeenat_fullname}
    \bibliography{references}
}

\end{document}

%% file: preamble.tex
\usepackage[dvipsnames]{xcolor}

\usepackage{mathtools}
\usepackage{cite}
\usepackage{amsmath,amssymb,amsfonts}
\usepackage{algorithmic}
\usepackage{textcomp}
\usepackage{siunitx}
\usepackage{xcolor}
\usepackage{blindtext}
\usepackage{float}
\usepackage{graphicx}
\usepackage{tabularx}
\usepackage{subcaption}
\usepackage{mwe}
\usepackage{tablefootnote}
\usepackage[export]{adjustbox}
\usepackage{subfiles} 
\usepackage[font=footnotesize]{caption}
\usepackage{epsfig}
\usepackage{soul}

%% file: sec/0_abstract.tex
\begin{abstract}
Image-based methods to analyze food images have alleviated the user burden and biases associated with traditional methods. However, accurate portion estimation remains a major challenge due to the loss of 3D information in the 2D representation of foods captured by smartphone cameras or wearable devices. In this paper, we propose a new framework to estimate both food volume and energy from 2D images by leveraging the power of 3D food models and physical reference in the eating scene. Our method estimates the pose of the camera and the food object in the input image and recreates the eating occasion by rendering an image of a 3D model of the food with the estimated poses.
We also introduce a new dataset, SimpleFood45, which contains 2D images of 45 food items and associated annotations including food volume, weight, and energy. Our method achieves an average error of 31.10 kCal (17.67\%) on this dataset, outperforming existing portion estimation methods. The dataset can be accessed at: \href{https://huggingface.co/datasets/purdueviperlab/SimpleFood45}{SimpleFood45}\footnote{Dataset - https://huggingface.co/datasets/purdueviperlab/SimpleFood45} and the code can be accessed at: \href{https://gitlab.com/viper-purdue/monocular-food-volume-3d}{monocular-food-volume-3d}\footnote{Code - https://gitlab.com/viper-purdue/monocular-food-volume-3d}.
\end{abstract}

%% file: sec/1_intro.tex
\section{Introduction}
\label{sec:intro}

Dietary assessment is essential for understanding and promoting healthy eating habits, serving as a key indicator of an individual's health~~\cite{liese2015dietary, shao2021_ibdasystem}.
Traditional methods, such as the 24-hour recall, heavily rely on user-reported data, introducing inherent limitations and biases ~\cite{poslusna2009misreporting}. In recent years, the rise of image-based dietary methods has garnered attention for alleviating the user burden associated with traditional approaches while demonstrating high accuracy ~\cite{gao2018food, jia2014accuracy, yang2019image, he2020multitask, he2021end}.

However, a significant challenge emerges for methods relying solely on 2D images - the loss of crucial information when projecting a 3D food object onto a 2D image plane. This limitation has spurred investigations into multi-view images ~\cite{7792736, 9635418, xu2013image} 
and depth-based methods ~\cite{thames2021nutrition5k, lo2018food, 8695351}, aiming to capture richer information than a single image can provide.

The emergence of readily available 3D data mitigates the loss of information associated with 2D images.
Datasets such as ShapeNet~\cite{chang2015shapenet}, Omni-Object3D~\cite{wu2023omniobject3d} among others, have enriched the space of 3D data. The most recent work introduces 
NutritionVerse3D~\cite{Tai_Chen_Keller_Kerrigan_Nair_Pengcheng_Wong_2023}, which provides 3D representations of food objects, offers a crucial tool for addressing the lack of 3D information in publicly available food image datasets. 
However, there are currently no methods that use this emerging and readily available 3D data for single image portion estimation.

In this work, we introduce a new framework to estimate food portion size by harnessing the power of 3D models while capitalizing on the simplicity and abundance of 2D food images. 
The key premise of our method is to recreate and render the eating occasion in the 2D input image using the available 3D food models.
In 3D space, the key parameters to recreate the 2D input image are the position and orientation of the camera and the food(s) in the image. Our proposed method estimates these parameters and uses them to render an image of a 3D model corresponding to the food(s) in the image.
The known volume of the 3D model and a scaling factor, which is the ratio of the area occupied by the food(s) in the rendered image to the input image, are used to resize the 3D model by the scaling factor to produce the estimated volume of the food(s). The area occupied by the food(s) in the input image is obtained from a segmentation mask which is obtained from a neural network segmentation model.
From the estimated volume, the USDA Food and Nutrition Database for Dietary Studies (FNDDS)~\cite{MONTVILLE201399} is used to obtain the energy value. 

Unlike existing image-based methods reliant on neural networks \cite{9733557, thames2021nutrition5k, 9874714,foods12234293}, our proposed method does not rely on complex neural architectures for portion estimation. Only standard tasks such as food classification and segmentation are performed using neural network models while the portion estimation is based solely on the 3D geometry of the food and the estimated camera and object poses.

Many existing food portion estimation methods are evaluated on private datasets (Table~\ref{tab:1}), making it difficult for comparison. 
Further, there are no image datasets that contain ground-truth food volume and corresponding 3D models to evaluate our proposed method. 
Some existing works use the Nutrition5k dataset ~\cite{thames2021nutrition5k} which is a publicly available large-scale food dataset with real food images that contain nutritional information. Around 3,000 images in the dataset have an associated depth map which is a pixel-wise mapping of the distance from the camera to the objects in the image. However, the camera position is fixed and it only captures the top view of the foods. This significantly reduces the generalizability of this dataset since typical images of foods captured from smartphone cameras and wearable devices have various camera poses.
Therefore, we introduce a new dataset called  \textit{SimpleFood45} that comprises images of real foods captured from different camera poses with a checkerboard as the physical reference. The images are captured using a smartphone camera to simulate a typical eating occasion.
Furthermore, each image contains the ground-truth volume (milliliters), weight (grams), and energy (kCal). 
The dataset aims to provide a better baseline for the evaluation of food portion estimation methods.

Key contributions of our paper include the following. 1) We propose a lightweight, geometric-based framework that enables the utilization of existing 3D data for portion estimation from 2D images.  2) We introduce a new food image dataset, \textit{SimpleFood45}, which contains images captured using a smartphone camera with a physical reference and ground-truth volume for portion estimation. 3) Our method outperforms both neural network and 3D representation-based methods on the SimpleFood45 dataset. 

\begin{figure*}[ht!]
\centering
\includegraphics[width=\textwidth]{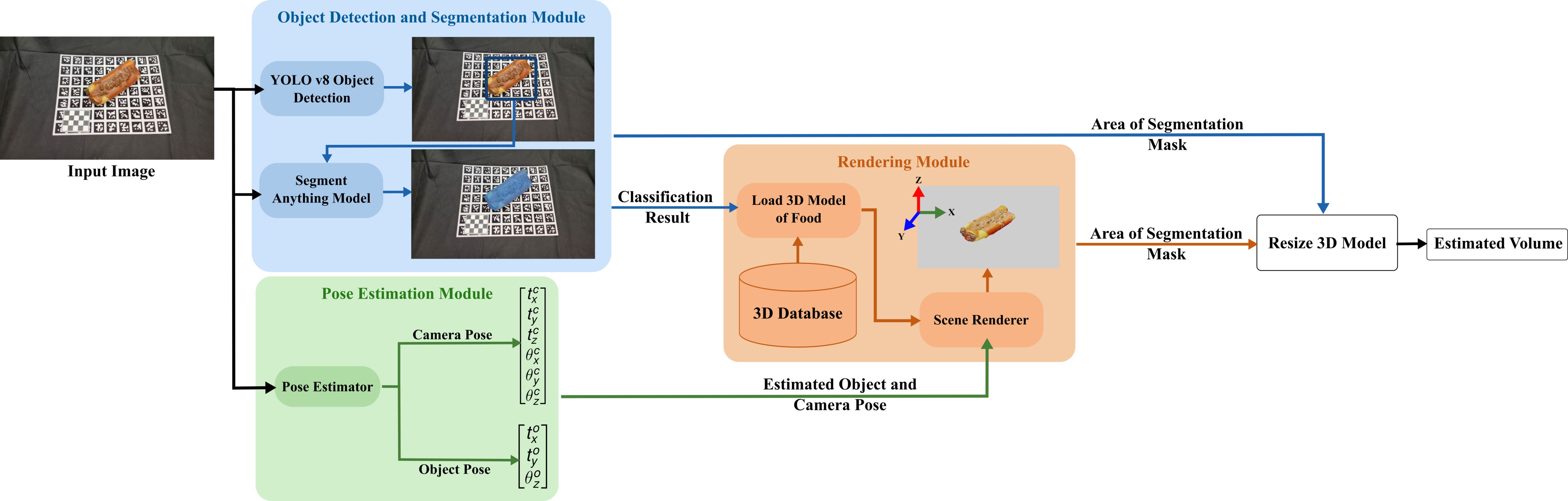}
\caption{\textbf{Overview of proposed method.} The system is divided into 3 modules, the Object Detection and Segmentation Module which uses the 2D image as an input and outputs a segmentation mask. The Pose Estimation Module estimates the camera pose and the orientation and translation of the food. The Rendering Module loads the 3D model based on the class label provided by the Object Detection and Segmentation Module and the pose parameters from the Pose Estimation Module to render an image of the food(s) in the input image. The size of the binary masks of the rendered image and the input image are compared to re-scale the 3D model to obtain the estimated volume.}
\label{fig:overview} \vspace{-0.4cm}
\end{figure*}

%% file: sec/2_related_works.tex
\section{Related Works}
\label{sec:relaterd_works}

Food portion estimation or volume analysis can be categorized into four main groups \cite{Lo2020}. 

\textbf{Stereo Based Approach.} These methods rely on multiple frames to reconstruct the 3D structure of the food. The food volume is estimated in \cite{5403087} from multi-view stereo reconstruction relying on epipolar geometry. Two-view dense reconstruction is performed in \cite{6395199}. 
Simultaneous Localization And Mapping (SLAM) is used in \cite{8329671} for continuous and real-time food volume estimation.
The main limitation of these methods is the requirement of multiple images for volume estimation which is not practical for real-world deployment.

\textbf{Model Based Approach.} These methods use predefined shapes and templates to estimate target volume. Model-based approaches are used in \cite{6738522} where certain templates are assigned to foods from a library and transformations based on some physical references are used to estimate the size and location of the food. A similar template matching approach is used in \cite{jia20123d} to estimate food volume from a single image. However, exact matching to these predefined templates cannot accommodate the variations of the foods. 

\begin{table}[ht]
\centering
\caption{Overview of datasets used in existing portion estimation methods. An asterisk (\textbf{*}) indicates datasets that are not publicly available.}
\begin{tabularx}{\columnwidth}{@{}X@{}}  
\toprule
\textbf{Dataset Description} \\
\midrule
\textbf{Stereo Based Approach} \\
Approx. 400 image sets\textbf{*} \cite{5403087} \\
\hdashline
Pair of images of six different fruits\textbf{*} \cite{6395199} \\
\hdashline
23 video segments with two types of food (iPhone 6+, wearable camera)* \cite{8329671} \\
\midrule
\textbf{Model Based Approach} \\
35 images per food item of a total of five food items\textbf{*} \cite{6738522} \\
\hdashline
30 images per replica of a total of seven replicas\textbf{*} \cite{jia20123d} \\
\midrule
\textbf{Depth Camera Based Approach} \\
3,000 images of 14 types of fruit replicas\textbf{*} \cite{8868629} \\
\hdashline
Nutrition5k dataset \cite{thames2021nutrition5k} \\
\midrule
\textbf{Deep Learning Approach} \\
192 dietary study images\textbf{*} \cite{9733557} \\
\hdashline
909 images from Nutrition5k \cite{9874714} \\
\hdashline
Detailed data from the Nutrition5k dataset \cite{thames2021nutrition5k} \\
\bottomrule
\end{tabularx}\vspace{-0.2cm}
\label{tab:1}
\end{table}

\textbf{Depth Camera Based Approach.} In this approach, a depth camera is used which produces a depth map capturing the distance of the camera to the foods in the image. The depth map is used in \cite{8868629, 10220023} to form a voxel representation of the image which is then used to estimate the volume of the food. The main limitation is the additional requirement of high-quality depth maps and additional post-processing needed for consumer depth sensors.

\textbf{Deep Learning Approach.} Neural network-based methods have utilized the abundance of image data for training complex networks to estimate food portion.
Regression networks are used in \cite{9733557, 9874714} to estimate the energy value of the food from a single image input and from an ``Energy Distribution Map" which is a pixel-to-pixel mapping between the input image and the distribution of energy of the foods in the image. Regression networks using the input image and depth maps are trained in \cite{thames2021nutrition5k} to produce the energy, mass, and macronutrient information of the food(s) in the input image.
Deep learning-based methods rely on large amounts of data to train the model and are generally not explainable. Their performance is often degraded when the input image is dissimilar from the training data.


The datasets used by various portion estimation methods detailed in Table~\ref{tab:1} highlights the gap in publicly available datasets for portion or energy estimation of foods. Additionally, non deep learning based methods rely on simple foods with rigid geometries. The Nutrition5k dataset \cite{thames2021nutrition5k} only provides top-view images that contain a physical reference (depth map) but do not contain ground-truth portion information. Therefore, there is a need for a public dataset that has a physical reference, ground-truth portion, and rigid foods for validation of portion estimation methods.

In this paper, we introduce a new perspective to the problem of food portion estimation by using 3D food models but relying solely on 2D food images as an input. We also introduce the SimpleFood45 dataset for the portion estimation benchmark.

%% file: sec/3_method.tex
\section{Method}
We propose an end-to-end framework that takes an input 2D image, estimates the camera pose and pose of the food(s) in the image, and renders an image of a 3D model of the food(s) using the estimated poses. The relative difference in the area occupied between the food(s) in the input image and the rendered image is utilized to scale the known volume of the 3D model to obtain the volume estimate. This 3D model is obtained by scanning real food items using a 3D scanner. One 3D model of each food type in the SimpleFood45 is collected and forms our 3D database. Our framework is illustrated in Figure~\ref{fig:overview}. 

This proposed method is divided into 3 modules. The \textit{Object Detection and Segmentation Module} is responsible for the classification and segmentation which provides the area occupied by the food(s) in the input image. The \textit{Pose Estimation Module} estimates the location and orientation of both the camera and the food(s) in the 3D world coordinates which are essential for rendering a 3D model.  
Finally, the \textit{rendering module} uses the 3D model from the 3D database based on the food classification result and the estimated poses for image rendering.
The ratio of the area occupied by the food(s) in the rendered image as compared to the area occupied by the food(s) in the input image is used to scale the 3D model and find the estimated volume of the food in the image. 
\begin{figure*}[t]
    \centering
    \begin{subfigure}[b]{0.225\textwidth}
        \centering
        \includegraphics[width=0.99\textwidth, frame]{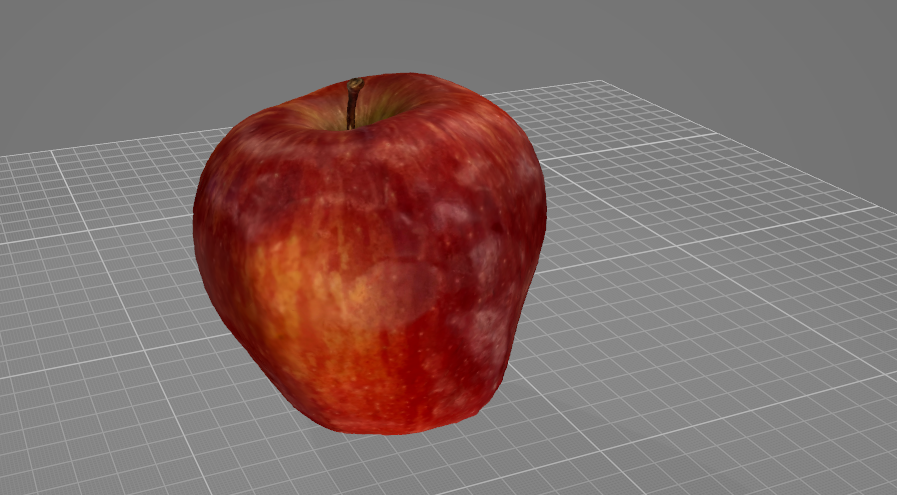}
        \caption{3D Model of Food}
        \label{fig:3d_model}
    \end{subfigure}
    \hfill
    \begin{subfigure}[b]{0.225\textwidth}  
        \centering 
        \includegraphics[width=0.99\textwidth, frame]{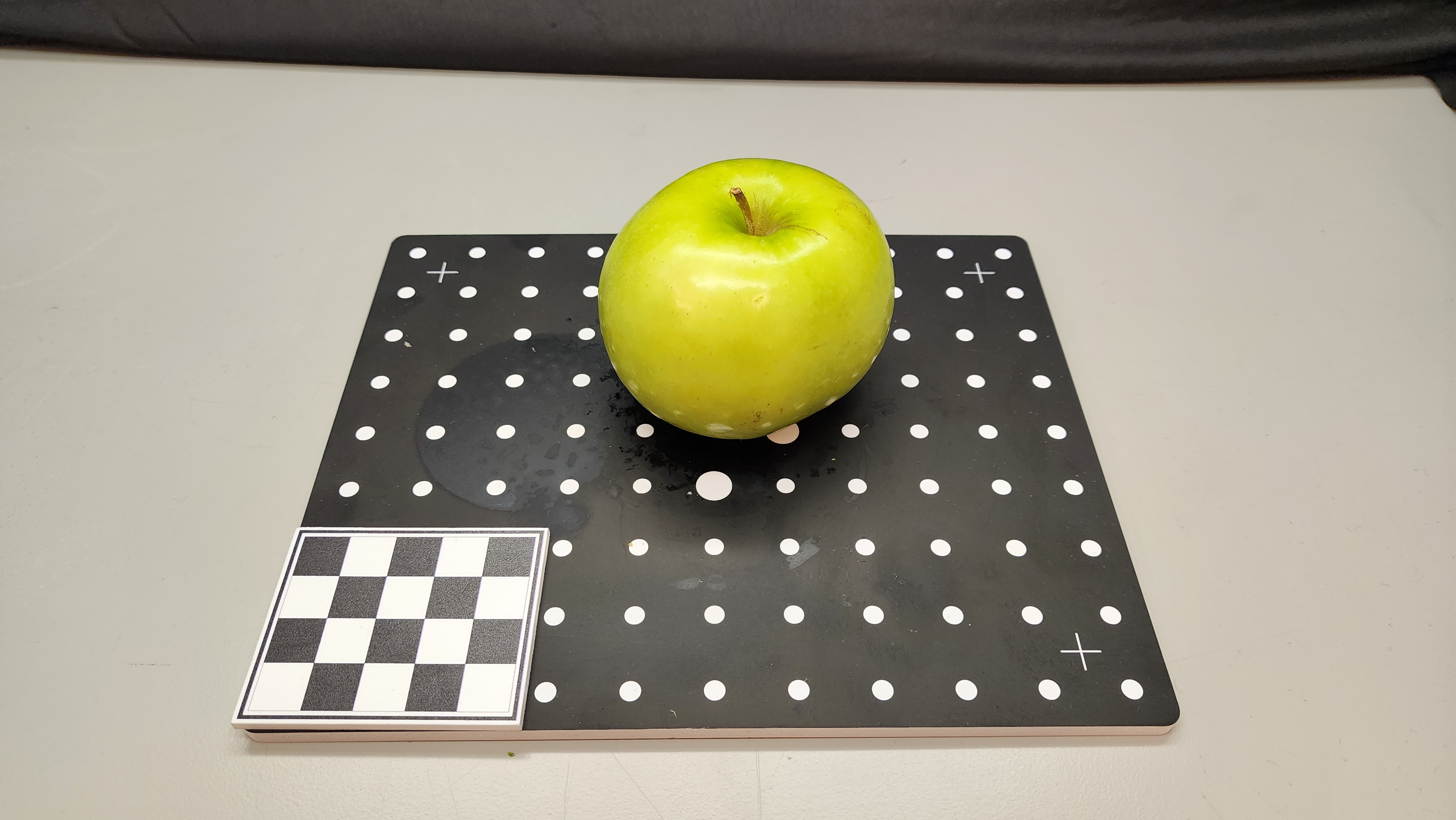}
        \caption{Input Inference Image}
        \label{fig:original_image}
    \end{subfigure}
    \hfill
    \begin{subfigure}[b]{0.225\textwidth}   
        \centering 
        \includegraphics[width=0.99\linewidth, frame]{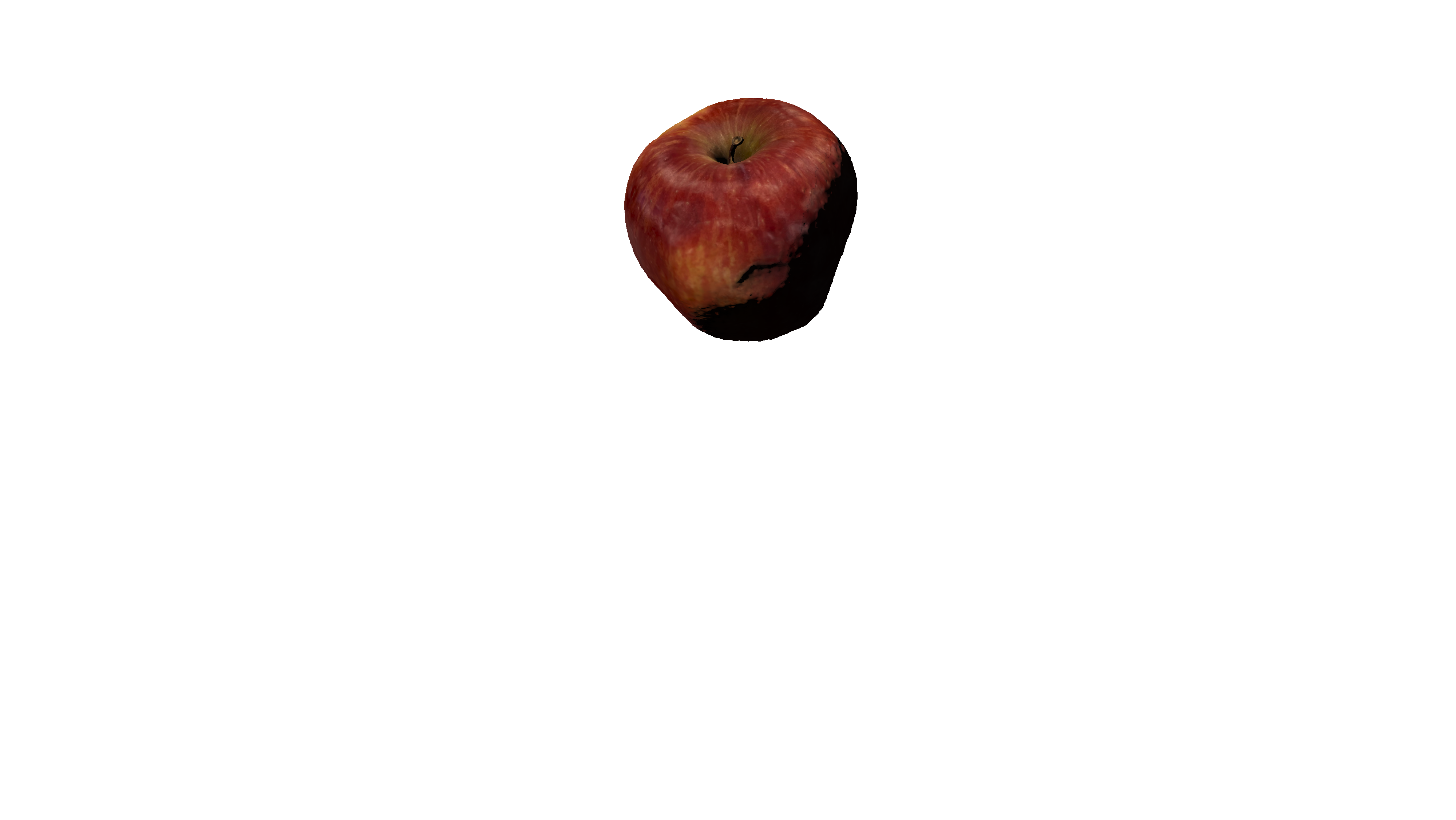}
        \caption{Rendered Image (Visualized)}
        \label{fig:rendered_texture}
    \end{subfigure}
    \hfill
    \begin{subfigure}[b]{0.225\textwidth}   
        \centering 
        \includegraphics[width=0.99\textwidth, frame]{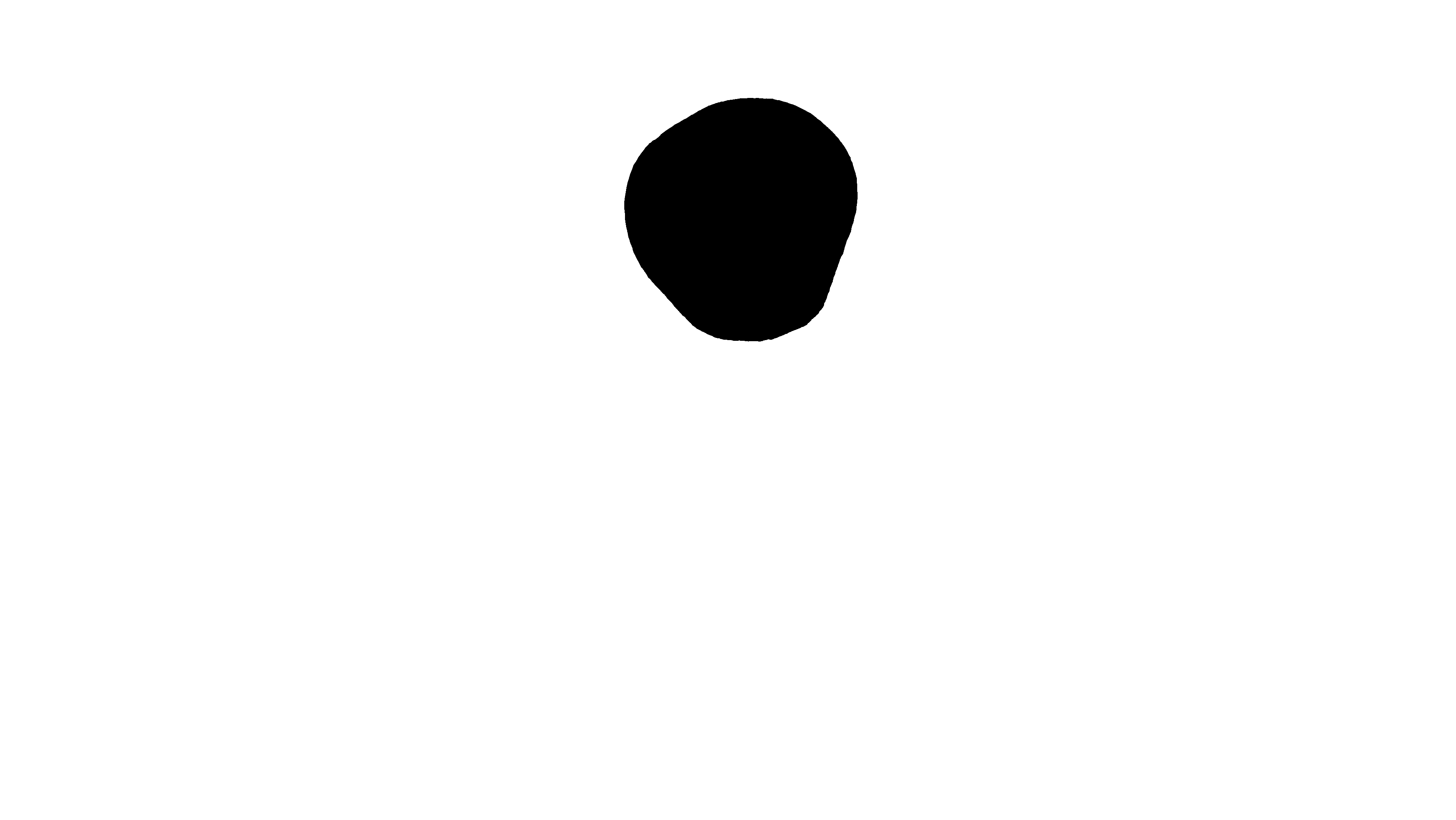}
        \caption{Rendered Image Mask}
        \label{fig:rendered_mask}
    \end{subfigure}
    \caption{The result of rendering an image of an apple based on the estimated pose of the input image using the 3D model of the apple.}  \vspace{-0.4cm}
    \label{fig:rendering}
\end{figure*}

\subsection{Object Detection and Segmentation Module} \label{sec:segmentation}
The objective of this module is to obtain a segmentation mask and classification label for the food(s) in the image. 
The classification label will determine the food type so that the corresponding 3D model of that food type can be used for image rendering while the segmentation provides information about the area occupied by the food(s) in the image. 
The Segment Anything Model (SAM) \cite{kirillov2023segany} offers zero-shot generalization which eliminates the need to find a suitable dataset with segmentation masks and fine-grained food classes for training. 
To provide a bounding-box prompt to the SAM for more accurate segmentation, the YOLOv8 \cite{Jocher_YOLO_by_Ultralytics_2023}, which takes an input image and provides a class label with a corresponding bounding box for each food in the image, is used.
In this work, we leverage the pre-trained YOLOv8 network on the VIPER-FoodNet (VFN)dataset \cite{mao2021visual, mao2021_nutri_hierarchy, he2022long}, which contains the most common and frequently consumed food types in the United States of America. Specifically, an input image is passed through the YOLOv8 network to obtain a class label and bounding box which is then used as a prompt to segment the input image using the SAM. Thus, we obtain a segmentation mask associated with each class label.

\subsection{Pose Estimation Module} \label{sec:pose_estimation}

\subsubsection{\textbf{Camera Pose Estimation}} \label{subsection:PnP}
The objective of camera pose estimation is to find the orientation of the camera in 3D world coordinates given the intrinsic camera matrix $\mathbf{K}$, which is obtained from camera calibration \cite{888718}, and the 2D pixel coordinates to 3D world coordinate correspondences of the 12 corner points of the checkerboard. 

The mapping from 3D to 2D for the $k$-th correspondence is described by:
\begin{equation} \label{eq:1}
\begin{bmatrix}u_k \\ v_k \\ 1\end{bmatrix} = \mathbf{K}_{3 \times 3}\left[\mathbf{R}|\vec{\mathbf{t}}\right]_{3 \times 4} \begin{bmatrix}
    X_k \\ Y_k \\ Z_k \\ 1
\end{bmatrix}
\end{equation}
where $(u_k, v_k)$ are the pixel coordinates of the $k$-th corner point and $(X_k, Y_k, Z_k)$ is its corresponding location in 3D world coordinates. 
The values of $(X_k, Y_k, Z_k)$ are chosen arbitrarily with the constraint that the 12 points of the checkerboard form a uniform grid pattern in 3D space.
The rotation and translation of the camera in 3D world coordinates are represented by a rotation matrix $\mathbf{R}_{3 \times 3}$ and a translation vector $\vec{\mathbf{t}}_{3 \times 1}$.

Equation~\ref{eq:1} cannot be solved directly because the correspondences picked out often result in a rank-deficient matrix \cite{hartley2003multiple}. Therefore, the Perspective-n-Point algorithm (PnP) \cite{lepetit2009ep} is employed which solves for a rotation matrix $\mathbf{R_c}$ and translation vector $\vec{\mathbf{t}}_\mathbf{c}$ by making it an optimization problem that minimizes the reprojection loss of Equation~\ref{eq:1}. However, The extrinsic camera matrix obtained from the PnP,  $[\mathbf{R_c}|\vec{\mathbf{t}}_\mathbf{c}]$, is a mapping from the world coordinate system to a camera coordinate system where the origin of this camera coordinate system is the camera itself. Given a point $\vec{\mathbf{X}}_\mathbf{c}$ in the camera coordinate system and a corresponding point $\vec{\mathbf{X}}_\mathbf{w}$ in the 3D world coordinates, their relation is described by:
\begin{equation}\label{eq:2}
    \vec{\mathbf{X}}_\mathbf{w} = \left[ \mathbf{R_c|}\vec{\mathbf{t}}_\mathbf{c} \right]\vec{\mathbf{X}}_\mathbf{c} = \mathbf{R_c}\vec{\mathbf{X}}_\mathbf{c} + \vec{\mathbf{t}}_\mathbf{c}
\end{equation}

We need to map the camera coordinate system to the 3D world coordinate system to determine the camera's position and angle relative to the origin of the 3D world coordinate system.
To obtain this mapping, we determine the origin $\vec{\mathbf{0}}$ of the 3D world coordinate system to the camera coordinate system using Equation~\ref{eq:2}:
\begin{equation}
\begin{split}
    \vec{\mathbf{0}} = &\mathbf{R_c} \vec{\mathbf{X}}_\mathbf{c} + \vec{\mathbf{t}}_\mathbf{c} \\
    \implies & \vec{\mathbf{X}}_\mathbf{c} = -\mathbf{R_c}^{-1}\vec{\mathbf{t}}_\mathbf{c}
\end{split}
\end{equation}
Here, $ \vec{\mathbf{X}}_\mathbf{c}$ will be the position of the camera relative to the origin of the 3D world coordinate system, and the rotation relative to the 3D world coordinate system would be the inverse of $\mathbf{R_c}$.
Since rotation matrices by definition are orthogonal matrices, it is known that $\mathbf{R_c}^{-1} = \mathbf{R_c'}$. 
Therefore, we can obtain the camera pose in the 3D world coordinate system through:
\begin{equation} \label{eq:5}
\begin{split}
    \mathbf{R} = \mathbf{R_c}^{-1} = \mathbf{R_c'} \\
    \vec{\mathbf{X}}_\mathbf{c} = -\mathbf{R_c'}\vec{\mathbf{t}}_\mathbf{c} = \vec{\mathbf{t}}
\end{split}
\end{equation}
where $\mathbf{R}$ and $\vec{\mathbf{t}}$ are the camera's rotation and translation in the 3D world coordinates which constitutes the camera pose.



\subsubsection{\textbf{Object Pose Estimation}}
The position and orientation of the food(s) in the input image in the 3D world coordinate system are estimated using the segmentation mask and checkerboard pattern. 
To estimate the object orientation, the pixel coordinates of the points in the segmentation mask of the input image are taken as points in 2D space. We perform Principal Component Analysis (PCA) \cite{pearson1901liii} on these points. The eigenvector (principal component) corresponding to the largest eigenvalue gives us the direction of orientation of the major (longer/larger) part of the food. The angle $\theta^o_z$ that the largest principal component makes with the horizontal axis is an approximation of the orientation of the object. The 3D world coordinate is defined with the $Z$-axis pointing upwards so this estimated orientation $\theta^o_z$ is the object's rotation about the $Z$-axis. Our proposed method does not estimate the rotation of the object along the $X$ and $Y$ directions and these values $\theta^o_x$ and $\theta^o_y$ are assumed to be $0$. 



Next, we leverage the checkerboard pattern to estimate the object's position. 
The object needs to be translated only on the $XY$-plane of the 3D world coordinate system since the food is usually kept on the same surface as the checkerboard ($Z$ translation is 0 relative to the checkerboard). 
The corners of the checkerboard form a square grid that is $1.2$ cm apart, image rectification using DLT \cite{hartley2003multiple} is performed.
In the rectified image, the scale from pixel to cm is known. The distance between the top-right corner of the checkerboard and the center of the segmentation mask of the food (assumed to be the center of the food) is estimated from this rectified image. This results in the translation along $X$ and $Y$ axes as $t^o_x$ and $t^o_y$ respectively.
From the approximated rotation and translation values, the object pose is then $\mathbf{O} = \left[ t^o_x, t^o_y, \theta^o_z \right]$.

\subsection{Rendering Module}
The mapping from the pixel coordinate system to the 3D world coordinate system is obtained from the Pose Estimation Module. The class label obtained from the YOLOv8 network is used to load the corresponding 3D model of the food from the 3D database. In the rendering module, the 3D world coordinate system is mapped back to the image coordinate system to render an image from the 3D model.  

The object is translated and oriented in 3D space according to the object pose $\mathbf{O}$.
The estimated camera pose (Equation~\ref{eq:5}) and the camera intrinsic matrix $\mathbf{K}$ together form the camera projection matrix $\mathbf{P}_{3 \times 4} = \mathbf{K}_{3 \times 3} \left[ \mathbf{R|\vec{\mathbf{t}}}\right]_{3 \times 4}$
We obtain the pixel location $\mathbf{x}_i$ in homogeneous coordinates of each 3D point $\mathbf{X}_i$ of the 3D object using this projection matrix
\begin{equation}
    \mathbf{x}_i = \mathbf{P}_{3 \times 4}\mathbf{X'}_i \: \: \: \forall i
\end{equation}
Finally, a rendered scene using the estimated camera pose of the input image and the estimated object pose is obtained. 

Our proposed method requires only a binary mask of the object in the image, therefore the 3D object is rendered without any texture (color).
Figure~\ref{fig:rendering} illustrates the input and output of the rendering module using an input image containing an apple (Figure~\ref{fig:original_image}). The 3D model of an apple (Figure~\ref{fig:3d_model}) is used to render an image using the estimated camera and object pose from Figure~\ref{fig:original_image}. For visualization, Figure~\ref{fig:rendered_texture} shows the rendered image with its texture, but Figure~\ref{fig:rendered_mask} is the actual rendered binary image mask used for volume estimation. The outputs Figure~\ref{fig:rendered_texture} and Figure~\ref{fig:rendered_mask} may not exactly match the input image Figure~\ref{fig:original_image} because the output camera and object poses may contain pose estimation errors.

\subsection{Volume Estimation}
The rendered image recreates the input image using the 3D model of the food type in the image. Therefore, we can assume that the ratio of the areas occupied by the food(s) in the rendered image and the input image is proportional to the ratio of the volumes of the objects in these images.
The area occupied by the food is estimated by the number of pixels in the segmentation mask of that food.
\begin{equation}
    A = \sum_{p \in S} 1
\end{equation}
where $p$ denotes each pixel belonging to the segmentation mask $S$ of the input image. Similarly, $A'$ is the area occupied by the object in the rendered image:
\begin{equation}
    A' = \sum_{p \in S'} 1
\end{equation}
where $p$ denotes each pixel belonging to the segmentation mask $S'$ of the rendered image. 

The 3D model is scaled by a factor $s$:
\begin{equation}
    s =\sqrt{A / A'} 
\end{equation}
which is the one-dimensional ratio of the areas occupied by the objects in the input and the rendered images. 
The estimated volume $\tilde{v}$ of the object in the input image is approximated as the volume of this scaled 3D point cloud. 
The energy-to-volume ratio (energy density) is obtained from the FNDDS database \cite{MONTVILLE201399} by finding the ratio of the energy (kCal) to the volume (mL) of a standard serving size for the corresponding food type. This energy density $\rho$ is multiplied by the estimated volume to obtain the estimated energy of the food(s) in the input image:
\begin{equation}
    \tilde{e} = \rho \tilde{v}
\end{equation}

\subsection{SimpleFood45 Dataset Collection}

\begin{figure}[h!]
    \centering
    \includegraphics[width=\columnwidth]{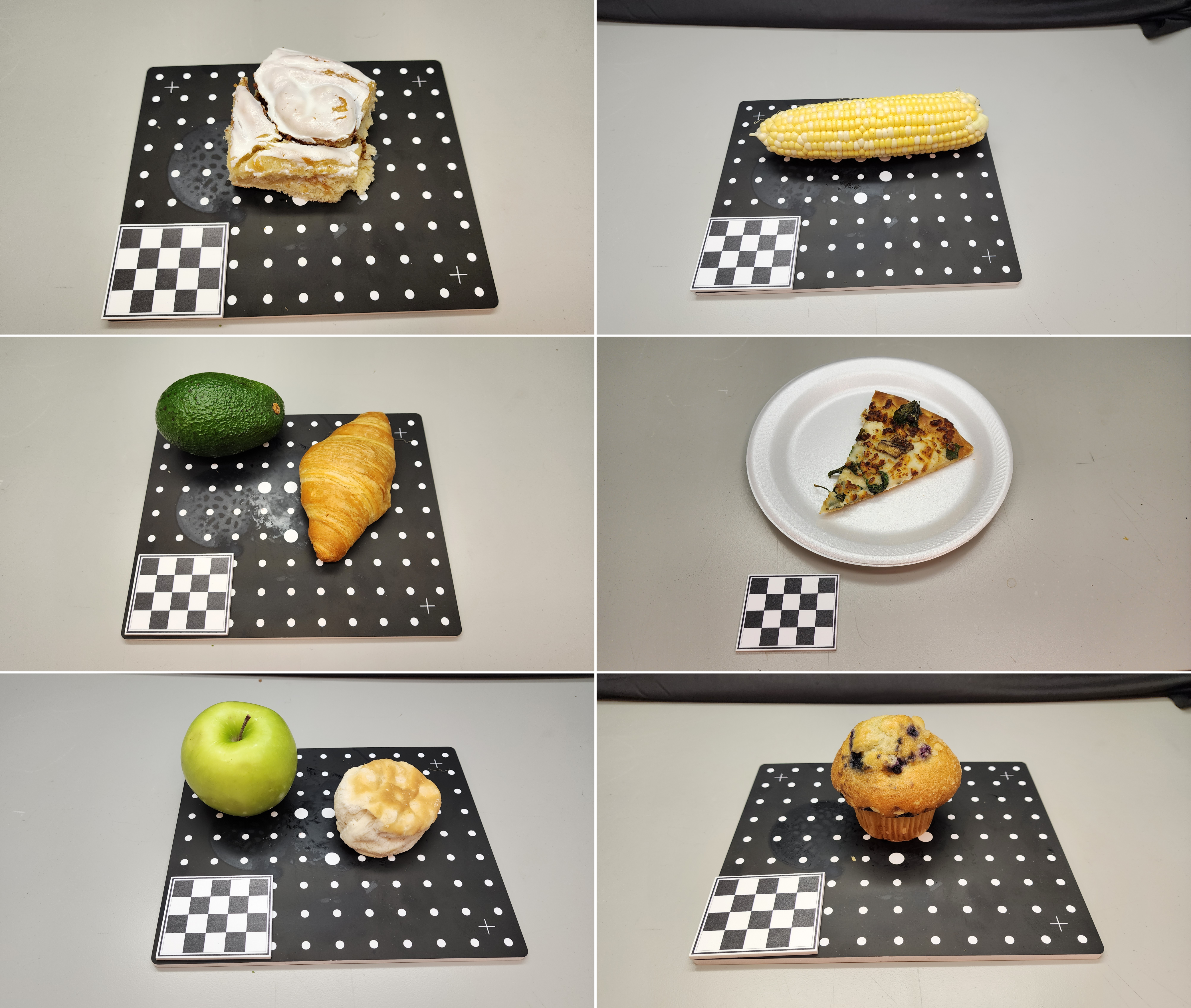}
    \caption{Image samples from the SimpleFood45 dataset. The samples feature different food types and different camera and object poses}. \vspace{-0.4cm}
    \label{fig:sample}
\end{figure}

The SimpleFood45 contains 12 food types and a total of 513 images of real foods with ground-truth class label, volume(mL), weight(g), and energy(kCal).  The images are captured using a Samsung Galaxy S22 Ultra smartphone. Each image contains a checkerboard grid of size $5 \times 4$ yielding a total of $4 \times 3$ inner corner points. The physical distance between each corner point is $1.2$cm. Figure~\ref{fig:sample} shows some examples of the images in SimpleFood45 dataset.

A Revopoint POP2 \cite{Revopoint} 3D Scanner is used to collect the 3D models of the foods.
Each of the 12 food types has one 3D scanned model, but 3-4 corresponding food items. Here, the food type refers to a general category of food (Eg. Apple) while the food item refers to an instance within the food type. 
Each food item has at least 10 images and the camera pose and the pose of the food item are different for each image. 
The volume occupied by the 3D model in 3D space is taken as the ground-truth volume since the physical dimensions of the 3D model and the actual food are the same.Further, the FNDDS database \cite{MONTVILLE201399} is used to obtain the energy (kilocalories) for each food item using the ground-truth weight and the food type matching. 

%% file: sec/4_experimental_results.tex
\section{Experimental Results}

\begin{table*}[ht]
  \centering
  \footnotesize
   \caption{\textbf{Comparison with other methods}. The neural network-based methods are made to directly regress the calorific value of the food and hence do not have any metrics for volume output. The 3D Assisted Portion Estimation outperforms all other methods for energy estimation.} 
  \begin{tabular}{| c | c | c | c | c |}
  \hline
  \textbf{Method} & \textbf{VMAE (mL)} $\downarrow$ & \textbf{VMAPE (\%)} $\downarrow$ & 
  \textbf{EMAE (kCal)} $\downarrow$ & \textbf{EMAPE (\%)} $\downarrow$ \\
  \hline
    Baseline & 83.28 & 170.37 & 120.09 & 547.34 \\ 
    \hline
    \multicolumn{5}{|c|}{\textbf{Neural Network Based Methods}}\\
    \hline
    RGB Only \cite{9733557} & - & - & 273.56 & 222.72 \\
    Distribution Map Only \cite{9733557} & - & - & 216.73 & 159.48 \\
    Distribution Map Summing \cite{ma2023improved} & - & - & 192.76 & 93.16 \\
    2D Direct Prediction \cite{thames2021nutrition5k} & - & - & 195.28 & 351.50\\
    Depth as 4th Channel \cite{thames2021nutrition5k} & - & - & 173.63 & 182.29 \\
    \hline
    \multicolumn{5}{|c|}{\textbf{3D Representation Based Methods}}\\
    \hline
    Voxel Representation Estimation Per Food (VREPF) \cite{10220023} & \textbf{22.35} & 24.51 & 32.01 & 25.13 \\
    \textbf{3D Assisted Portion Estimation (Ours)} & 24.51 & \textbf{14.01} & \textbf{31.10} & \textbf{17.67} \\
    \hline
  \end{tabular}
  \label{tab:2}
\end{table*}

\subsection{Comparison With Other Methods}
Our proposed method is validated on the SimpleFood45 dataset. The performance of the proposed method is compared with existing neural network-based methods that have shown high accuracy \cite{9733557, ma2023improved, thames2021nutrition5k}. Since our proposed method uses 3D representation of food data, we also compare it against a voxel-based method that is based on \cite{10220023}.
The dataset is split into training and testing data using an 80-20 split ensuring equal representation of classes across the training and testing split.
All methods are evaluated only on the testing data to ensure fairness of comparison with the neural network based methods.

We use standard metrics including Mean Absolute Error (MAE) and Mean Absolute Percentage Error (MAPE) for comparison, and they are defined as follows:
\begin{equation}
\begin{split}
    \textbf{MAE  } = \frac{1}{N}\sum_{i=1}^N |(\hat{v}_i - v_i)| \\
    \textbf{MAPE (\%)  } = \frac{1}{N}\sum_{i=1}^N \frac{|(\hat{v}_i - v_i)|}{v}
\end{split}
\end{equation}
where $v_i$ is the ground-truth value, $\hat{v}_i$ is the estimated value of the $i$-th image, and $N$ is the number of images in the dataset. The volume estimation error is denoted by VMAE and VMAPE while the energy estimation error is denoted by EMAE and EMAPE. The units for VMAE and EMAE are mL and kCal, respectively.

The following methods \cite{9733557, ma2023improved, thames2021nutrition5k, 10220023} are used for comparison with our proposed method using the SimpleFood45 dataset since these methods have achieved portion estimation results with low error. Results are shown in Table~\ref{tab:2}.

\textbf{Baseline.} This is a common baseline used in regression tasks and also used in \cite{thames2021nutrition5k}. It is simply a model that predicts the mean ground-truth value of the dataset (volume/energy) for every instance $i=\{1,\dots,N\}$ of the dataset. For the volume baseline let the ground-truth volume of the $i$-th image be $v_i$. The mean ground-truth volume of the dataset is then 
\begin{equation*}
    \bar{v} = \sum_{i=1}^N v_i
\end{equation*}

The estimated volume $\Tilde{v}_i$ for the $i$-th image is then given by:
\begin{equation*}
    \Tilde{v}_i = \bar{v}
\end{equation*}

Similarly the energy baseline is the mean energy value of the dataset predicted for each iamge in the dataset. 

\textbf{Deep learning Based Methods.}
\textit{RGB Only:} a Resnet50 backbone and 2 fully connected layers are trained on the SimpleFood45 to regress the energy value of the input image \cite{9733557}.
\textit{Distribution Map Only:} The ground-truth ``Energy Density Map" \cite{9733557} is used as the input to directly regress the energy values of the foods.
\textit{Distribution Map Summing:} The ``Energy Distribution Maps" are summed up directly \cite{ma2023improved} instead of using a regression network. 
\textit{2D Direct Prediction:} A regression network that takes the input image and regresses the energy value with the same architecture as \cite{thames2021nutrition5k}.
\textit{Depth as 4th Channel:} First, the ZoeDepth Depth Estimation network \cite{bhat2023zoedepth} is used to generate a depth map for each image in the SimpleFood45 dataset. This depth map is appended to the RGB input image as the 4th channel and that is fed to the same backbone as 2D direct prediction \cite{thames2021nutrition5k}.

\textbf{3D Representation Based Methods.}
\textit{Voxel Representation Estimation Per Food (VREPF):} A voxel representation is created from the input image and the corresponding depth maps \cite{10220023}. Finally, the number of voxels occupied by a food is translated into actual physical units. Let the number of voxels for the $k$-th image belonging to the $i$-th food type be $\Tilde{V}_k$ and $V_k$ is its ground-truth volume. A scaling factor $$\rho^i_k = V_k / \Tilde{V_k}$$ is calculated for each image belonging to the $i$-th food type. The mean scaling factor $$\Bar{\rho}^i = (1/N) \sum^N_{k=1}\rho^i_k$$ where $N$ is the number of images in the dataset belonging to the $i$-th food type, is used to convert the voxel volume to an estimated volume. Therefore, the estimated volume is given by $$\hat{V}_k = \Tilde{V}_k \Bar{\rho}^i$$.

Our proposed method achieves significantly better results than the neural network based methods and outperforms the voxel representation based methods in most cases. Though the neural network models are trained on the SimpleFood45, they fail to accurately estimate the energy values of the food. This could be attributed to the limited data and the large variance in the energy values of the food types, which also speaks to the efficacy of our method. 

\subsection{Generalization to Other Datasets}
The proposed method requires food images with a physical reference and a corresponding 3D model of the food type for volume estimation. Currently, no publicly available dataset contains the required information to be used directly for comparison. 
Therefore, we select 3 compatible food types from the Nutrition5k dataset \cite{thames2021nutrition5k} which overlaps with food types from the SimpleFood45 dataset - Apple, Bagel, and Pizza. In the Nutrition5k dataset, the camera position is fixed for the top-view images and the known distance of the camera to the plate provides a physical reference for our method. The camera parameters are estimated for the camera model in \cite{thames2021nutrition5k}. 

\begin{table}[h]
  \centering
  \footnotesize
  \caption{\textbf{Results on Other Data}. The table shows results on the Nutrtion5k images using our proposed method. } 
  \begin{tabular}{| c | c | c | c |}
  \hline
  \textbf{Food}  & \textbf{Method}  & \textbf{EMAE (kCal)} $\downarrow$ & \textbf{EMAPE (\%)} $\downarrow$ \\
  \hline
    Apple & VREPF~\cite{10220023} & 9.38 & 12.36 \\
    Apple &  Ours & 11.5 & 14.85 \\ 
    \textbf{Apple} & \textbf{Ours*} & \textbf{8.68} & \textbf{ 11.68} \\
    \hline
    Bagel & VREPF~\cite{10220023} & 149.48 & 62.23 \\
    \textbf{Bagel} &  \textbf{Ours} & \textbf{35.74} & \textbf{14.79} \\ 
    \hline
    Pizza & VREPF~\cite{10220023} & 155.34 & 68.01 \\
    \textbf{Pizza} &  \textbf{Ours} & \textbf{97.65} & \textbf{35.62} \\ 
    \hline
  \end{tabular}
  \label{tab:3}
  \footnotesize{\\ * indicates that the 3D model source is from the NutritionVerse dataset \cite{Tai_Chen_Keller_Kerrigan_Nair_Pengcheng_Wong_2023} }
\end{table}

Table~\ref{tab:3} shows the portion estimation results of our proposed method and the Voxel Representation Estimation Per Food (VREPF) on images from the Nutrtion5k dataset. The voxel representation for VREPF is constructed using the depth maps provided in the Nutrtion5k dataset.
The only overlap between the food types in NutrtionVerse \cite{Tai_Chen_Keller_Kerrigan_Nair_Pengcheng_Wong_2023} and the SimpleFood45 is Apple. Therefore, for the images of apples, our proposed method is also evaluated on the Nutrition5k images using a 3D model from the NutrtionVerse dataset. In this case, we do not use any of our data for portion estimation.
Our proposed method outperforms VREPF in all food types achieving a remarkably low error rate. The VREPF shows poor performance on some food types due to the poor quality of the depth map for these images which leads to errors in the voxel representation. Our proposed method is able to easily adapt to different datasets without any prior knowledge such as training data. While the testing is limited due to the low overlap of food types between the datasets, the lower error rate compared to the VREPF where the actual depth map is used, speaks to the generalizability of our method.

\subsection{Ablation Analysis}
To understand the importance of Object Pose Estimation on our proposed method, we conducted an ablative study on the SimpleFood45 dataset. We consider the object pose vector $\mathbf{O} = \left[ t^o_x \: \: t^o_y \: \: \theta^o_z\right]$ and analyze the impact when some or all of these values are set to 0 (the pose is not estimated). Setting $t^o_x = 0$ means the object will only be translated along the $Y$-axis, while setting $t^o_y=0$ means the object is only translated along the $X$-axis. Setting $\theta^o_z = 0$ means the 3D model is not rotated along the Z-axis.

\begin{table}[h]
  \centering
  \footnotesize
  \caption{\textbf{Ablation Analysis.} The impact of the 3D location and orientation of the object in the rendered image is analyzed. The parameter $t^o_y$ plays a vital role in the accuracy of volume estimation.} 
  \begin{tabular}{| c | c | c | c | c |}
  \hline
  $\mathbf{\left[ t^o_x \: \: t^o_y \: \: \theta^o_z\right]}$  & \textbf{VMAE} & \textbf{VMAPE}  & \textbf{EMAE} & \textbf{EMAPE}\\
  \hline
  $\left[ 0 \: \: t^o_y \: \: \theta^o_z\right]$ & \textbf{24.33} & 14.5 & \textbf{30.24} & 20.2 \\
  $\left[ t^o_x \: \: 0 \: \: \theta^o_z\right]$ & 67.37 & 34.4 & 84.33 & 39.66 \\
  $\left[ 0 \: \: 0 \: \: \theta^o_z\right]$ & 75.44 & 38.14 & 93.74 & 43.29 \\
  $\left[ t^o_x \: \: t^o_y \: \: 0\right]$ & 25.37 & 14.11 & 31.24 & 19.48 \\
  $\mathbf{\left[ t^o_x \: \: t^o_y \: \: \theta^o_z\right]}$ \textbf{(Ours)} & 24.51 & \textbf{14.01} & 31.10 & \textbf{17.67} \\
  \hline
  \end{tabular}
  \label{tab:4}
\end{table}

From Table~\ref{tab:4}, we observe that the translation in the $Y$-axis, \textit{i.e.}, $t^o_y=0$ plays a major role. This is because translation along the $Y$-axis moves the 3D object closer to or farther away from the camera and thus influences the object's size on the image. 
Setting $t^o_x = 0$ means that the object is not translated along the horizontal axis. Our method looks at the area occupied by the food in the rendered image and translating the object along the horizontal axis does not have a significant impact on the area occupied by the food. 
Finally, $\theta^o_z=0$ affects only non-symmetrical food items so its impact is restricted to certain food types. Symmetric food items such as apple and bagel can be rotated along the $Z$-axis but would appear the same in the rendered image due to their symmetry.
Therefore, the Object Pose Estimation does play a crucial role in the performance of our method with the lowest EMAPE achieved when all the parameters are taken into consideration.

\subsection{Discussion}
The major limitation of the proposed method is the requirement of strong prior knowledge (3D model) for each food type. Inconsistencies between the 3D model and the food item will yield poor estimates. For example, the 3D model for the avocado features a whole avocado, but when the input image contains a slice of an avocado would lead to erroneous results. There needs to be a good similarity between the input object structure and the 3D model structure as is the case for most geometric based models. However, Table~\ref{tab:4} shows that even for more general case data, the proposed method is able to fairly estimate the energy content of the foods and even outperforms existing methods. 
The limitation of 3D model requirement can also be overcome by collecting more 3D models so that the proposed method can be utilized for a lot more food items. The larger the number of models, the more accurate would be the portion estimation.

%% file: sec/5_conclusion.tex
\section{Conclusion}

We propose an approach that leverages the simplicity of data availability through 2D images while incorporating 3D data to overcome inherent limitations associated with the missing 3D information. Our method successfully bridges the gap between 2D food images and 3D models for portion size estimation. Notably, our method avoids heavy reliance on neural networks and the availability of training data.
We also introduce a SimpleFood45 dataset containing real food images for portion estimation. Experimental results show that our method outperforms both neural network based and 3D representation based methods, yet has the advantage of generalizability without any training data. This work also provides a solid foundation for future research in 3D food analysis. 

The major limitation of this work is the need for 3D models specific to each type of food. Future efforts will aim to minimize this dependency by developing methods for 3D reconstruction that use these models primarily as training data, rather than during the inference process. 